\documentclass{article} 

\usepackage{iclr2020_conference}
\usepackage{arxiv}

\usepackage{times}

\usepackage{amsmath,amsfonts,bm}

\def\eqref#1{equation~\ref{#1}}

\def\1{\bm{1}}

\DeclareMathAlphabet{\mathsfit}{\encodingdefault}{\sfdefault}{m}{sl}
\SetMathAlphabet{\mathsfit}{bold}{\encodingdefault}{\sfdefault}{bx}{n}

\usepackage{hyperref}
\usepackage{url}
\usepackage{multirow}

\title{Using LSTM to Translate French to Senegalese Local Languages: Wolof as a Case Study}

\author{Alla Lo$^\dagger$, Cheikh Bamba Dione$^\ddag$, Elhadji Mamadou Nguer$^\Box$, Sileye O. Ba$^\Diamond$, Moussa Lo $^\Box$\\
Universite Gaston Berger (UGB) $^\dagger$, Saint Louis, Senegal \\
University of Bergen$^\ddag$, Bergen, Norway \\
Universite Virtuelle du Senegal (UVS) $^\Box$, Dakar, Senegal\\
Dailymotion$^\Diamond$, Paris, France
}

\iclrfinalcopy 
\begin{document}

\maketitle
\vspace{-5mm}
\section{Problem statement}
\label{sec:intro}
Nowadays, internet applications are changing people's life all over the world. Applications such as Google-Map allow people to locate themselves. Facebook, Snapchat, Instagram allow people to be connected with their friends. Amazon, Alibaba allow people to do effortless online shopping. Netflix allows people to watch video contents. 
For smooth interaction, these applications require an understanding of internet dominant languages such as English, Spanish, Chinese, Arabic, French, etc. Others languages, spoken in Africa, such as Eastern Ethiopian Amharic, Southern African Swahili, Western African Fulani or Wolof can not be used with these applications. The main reason being that compared to English, for instance, many of these languages are spoken by small sized populations.

In machine learning, natural language processing models have been developed to extract relevant information from electronic documents on the internet. These models mostly address documents written in well-resource languages such as English, Mandarin, Spanish, Arabic, French, etc. Apart from rare cases, low-resource language, such as Eastern Ethiopian Amharic, Southern African Swahili, Western African Fulani or Wolof, are seldom addressed.  Having machine learning models such as machine translation systems that are able to transform information from high to low resource languages would have great impact. This will allow population speaking low-resource languages to make full use of internet contents and applications.

In this paper, we present investigations we conduct toward developing a long short-term memory (LSTM) neural network machine translation model to translate French into  Wolof \citep{Hochreiter-Neuralcomputation-1997,Sutskever-NIPS-2014}. Up to our knowledge, this is the first time such task is being addressed.  First, we gathered a dataset of seventy thousand aligned French-Wolof sentences. Then, we trained LSTMs (bi-directional networks, and bi-directional LSTM with attention) to translate French sentences into Wolof. Evaluation of our models using BLEU scores show promising results. In the remainder of this article, we give details about the corpus, the models, and conducted experimental validation.

\section{A Corpus for French to Wolof Translation}
\label{sec:dataset}
Wolof language is mainly spoken in West Africa \citep{Gamble-1977}. It's usage is mainly centered around Senegal and it's neighboring countries such as Gambia, Mauritania, Mali, and Guinea. 
Historically, Wolof is the language of the Jolof Kingdom which was located inside the modern Senegambian region. 
Approximately, Wolof is spoken by about 10 million people. 
In Senegal, Gambia, and Mauritania, Wolof has an official language status, together with French, English, and Arabic.
Because of France's historical influence in Senegal, and also because Wolof and French are two Senegalese official languages, bilingual French-Wolof documents are available.

The French-Wolof parallel corpus is constructed in the following steps. First, appropriate corpus data sources (pdf, text,
html and doc) are identified. Then, text content is extracted from the sources.
Next, raw text data are split into monolingual and
bilingual texts. Finally, monolingual texts are translated from French into Wolof. Likewise, bilingual texts are manually corrected. The output
of this stage consists of bilingual (or multilingual) parallel
documents. During the data collection process, special emphasis was placed on the
quality of the content and translation.

One of the most challenging tasks for creating a parallel
corpus is sentence alignment. This consists in extracting from parallel French and Wolof corpora sentence pairs that are translations of one another. In this work, we used semi-automatic methods to align
Wolof and French texts at the sentence level. First, using
Python scripts, we extracted the bilingual texts contained
in all documents. Then, paragraphs of texts are split into sentences using a script-based sentence-splitter.
Three widely used open-source
tools were benchmarked: hunalign \citep{Varga-Hunalign-2007}, yasa \citep{Lamraoui-MTS-2013} and champollion \citep{Ma-LREC-2006}. hunalign is a
hybrid algorithm that combines the dictionary and length based methods. In contrast, yasa and champollion use
lexical-based approaches. The final results indicated
that hunalign achieves satisfactory performance.

As showed by Table \ref{tab:stats}, the final corpus content can be categorized in six major
domains : education, general, laws, legend, religion, and society. Corpus content statistics are showed in Table \ref{tab:stats}. \textit{Religion} contains the most sentences, followed by \textit{General}. The smallest
portion are \textit{Society} and \textit{Laws}. The final corpus comprises 71034 phrases. \cite{Nguer-LREC-2020} provides more details about the corpus content.

\begin{table*}[t]
\centering
\renewcommand{\arraystretch}{1.4}
\begin{small}
\begin{tabular}{llllll}\hline
Domains & Languages & Tokens & Average Length & Vocabulary & Sentences \\ \hline
\multirow{2}{*}{Education} & French & 36467  &7.0129 & 7603 & \multirow{2}{*}{5200}\\
& Wolof & 27869 & 5.3594 & 6597 &  \\ \hline
\multirow{2}{*}{Religion} & French & 831972  &23.5040 & 49764 & \multirow{2}{*}{35397}\\
& Wolof & 739375 & 20.8881 & 44301 &  \\ \hline
\multirow{2}{*}{General} & French & 169666  &6.5915 & 15925 & \multirow{2}{*}{25740}\\
& Wolof & 161921 & 6.2906 & 10731 &  \\ \hline
\multirow{2}{*}{Laws} & French & 10016  &17.6028 & 2738 & \multirow{2}{*}{569}\\
& Wolof & 9951 & 17.4886 & 2461 &  \\ \hline
\multirow{2}{*}{Legend} & French & 27780  &12.8492 & 6460 & \multirow{2}{*}{2162}\\
& Wolof & 26051 & 12.0495 & 5292 &  \\ \hline
\multirow{2}{*}{Society} & French & 25391  &12.9151 & 6398 & \multirow{2}{*}{1966}\\
& Wolof & 26266 & 13.3601 & 5412 &  \\ \hline
\end{tabular}
\end{small}
\caption{Statistic summary of the French-Wolof parallel corpus}
\label{tab:stats}
\end{table*}

\section{LSTM Models For Translation}
\label{sec:models}

To assess the quality of the corpus we trained three word embedding models on the  Wolof monolingual corpus: a
continuous bag of words model (CBOW), a Skip-gram
model and a Global vector for word representation model
(GloVe) \citep{Mikolov-NIPS-2013,Pennington-EMNLP-2014}. Conducted evaluations about word analogy tasks show that constructed embedding capture words semantic relatedness despite the moderated corpus size \citep{Lo-LREC-2020}. 

In addition to developing word embedding models, we used the corpus to train and evaluate four LSTM based models to translate French sentences into their corresponding Wolof version: baseline LSTM, bidirectional LSTM, LSTM+attention, bidirectional LSTM+attention \citep{Hochreiter-Neuralcomputation-1997,Sutskever-NIPS-2014}.

For the translation task, we split the dataset into two disjoints sets of 50\% each (i.e.\ 34805 parallel sentences) for training and validation, respectively. The data split was done randomly. The sentences in the corpus have different lengths. Some are short (e.g.\, length 20-30 words), but there are very long ones ($>$ 100 words). We filter out sentence pairs whose lengths exceed 50 words and used padding to compensate for the empty slots in shorter sentences.

Across experiments, the following hyper-parameters are kept constant: number of LSTM units, embedding size, weight decay, dropout rate, shuffle size, batch size, learning rate, max gradient norm, optimizer, number of epochs and early stopping patience. We used a baseline encoder-decoder LSTMs with 300 cells and 128 dimensional word embeddings. All  models  are composed of a  single  LSTM  layer  with  
a dropout layer for the decoder, dropout rate and weight decay regularization parameters for both the encoder and decoder. It has to be noted that word embeddings are learned together with the LSTM networks.

Models are trained using Adam stochastic gradient descent with a learning rate set to $10^{-3}$ \citep{kingma-Adam-arXiv-2014}. We perform early stopping based on the validation set accuracy. Our shuffled mini-batch contains 128 training sentences.

\begin{table*} [t]
  \centering 
  \begin{small}
  \begin{tabular}{|l|c||c|c|c|c|} \hline
  \textbf{Our NMT models} & \textbf{Accuracy} & \textbf{BLEU 1} & \textbf{BLEU 2} & \textbf{BLEU 3} & \textbf{BLEU 4} \\ \hline
    LSTM & 56.69 & 26.36 & 44.24 & 47.46 & 39.73  \\ \hline
    LSTM + attention & 63.89 & 23.63 & 41.93 & 45.97 & 37.78 \\ \hline
   LSTM + bidirectional & 68.03 & 25.29 & 43.36 & 46.90 & 39.34 \\ \hline
   LSTM + bidirectional + attention & 72.27 & 23.49 & 41.81 & 45.90 &  38.65 \\ \hline 
\end{tabular}
\end{small}
  \caption{\label{tab:result1} Models BLEU scores (in percentage) on French-Wolof translation task.}
\end{table*}

The quality of our translations are evaluated by comparing the predictions and ground truth using BLEU \citep{Papineni-BLEU-2002}. We use cumulative BLEU scores \cite{Brownlee-MLM-2017} by calculating the weighted geometric mean of individual n-gram scores from 1 to 4 orders. The weights for BLEU-1 are (1.0, 0, 0, 0) or 100\% for 1-gram. BLEU-2 assigns a weight of 0.50 to each of the 1-gram and 2-gram. BLEU-3 assigns a weight of 0.33 to each of the 1-, 2- and 3-gram scores. The weights for the BLEU-4 are 0.25 for each of the 1-gram, 2-gram, 3-gram and 4-gram scores. 
For each model, we report cumulative BLEU scores (BLEU 1 to 4) combined with the smoothing function \textit{method4} as suggested by \citet{Chen-WSMT-2014}. This function assigns smaller smoothed counts to shorter translations as they tend to have inflated precision values due to having smaller denominators. 

Table \ref{tab:result1} shows that the best BLEU scores are achieved by the baseline unidirectional non-attentional model. Our experiments showed that adding bidirectional LSTMs and the attention mechanism did not necessarily improve BLEU scores.  The most likely explanation is the noticeable difference in the number of parameters. In other words, part of this improvement is likely due to the fact that the baseline model has fewer parameters and therefore require less training data. 
However, despite the lower BLEU scores, analysis of the results shows that the bidirectional attentional LSTM produces better qualitative translations with less repeated words.

\section{Conclusion}
\label{sec:conclusions}

Research presented in this paper are preliminary investigations we are conducting about analyzing Wolof language using deep neural networks. In the work, words were consider as basic token exploited LTSM networks for the translation task. Our aim is to make our corpora and models publicly available to the machine learning community to advance research about low-resource African languages. 

In the future, we plan to consider byte pair encoding (BPE) tokenization \citep{Sennrich-ACL-BPE-2016} to account for Wolof's morphological and compositional structure. We also plan to investigate the use of transformer models for the translation task as transformer have stated latest state of the art natural language processing performances \citep{Devlin-BERT-2019}.

\subsubsection*{Acknowledgments}
Authors thank the CEA MITIC of Universit\'e Gaston Berger in S\'en\'egal for partly funding this work.

\bibliography{iclr2020_conference}
\bibliographystyle{iclr2020_conference}

\end{document}